\documentclass[lettersize,journal]{IEEEtran}

\usepackage{xcolor}
\usepackage[]{hyperref}
\hypersetup{colorlinks,breaklinks,linkcolor=blue,urlcolor=blue,anchorcolor=blue,citecolor=blue}

\usepackage{amsmath,amsfonts}
\usepackage{algorithmic}
\usepackage{algorithm}
\usepackage{array}
\usepackage[caption=false,font=normalsize,labelfont=sf,textfont=sf]{subfig}
\usepackage{textcomp}
\usepackage{stfloats}
\usepackage{url}
\usepackage{verbatim}
\usepackage{graphicx}
\usepackage{adjustbox}
\usepackage{cite}
\usepackage{bm, amssymb, fixmath,dsfont}
\let\oldnl\nl
\newcommand{\nonl}{\renewcommand{\nl}{\let\nl\oldnl}}
\usepackage{color}
\usepackage{soul}
\usepackage{multirow}
\usepackage{booktabs}
\usepackage{bbm}
\usepackage{tikz}
\usepackage{rotating}

\usepackage{circledsteps}
\usepackage{physics}
\usepackage{tabularx, makecell, setspace}
\usepackage{circledsteps}
\usepackage{physics}
\usepackage{amsthm}

\usepackage{adjustbox}
\usepackage{svg}
\usetikzlibrary{shapes.geometric, arrows, positioning}
\usepackage{arydshln}

\title{\LARGE \bf
	Hybrid Differential Reward: Combining Temporal Difference and Action Gradients for Efficient Multi-Agent Reinforcement Learning in Cooperative Driving
}

\author{
	Ye Han, Lijun Zhang$^*$, Dejian Meng, Zhuang Zhang  
	\thanks{Ye Han, Lijun Zhang, Dejian Meng, Zhuang Zhang are with the School of Automotive Studies, Tongji University, Shanghai 201804, China.
		{\tt\small \{hanye\_leohancnjs, tjedu\_zhanglijun, mengdejian, zhuang\_zhang\}@tongji.edu.cn}}%
	\thanks{$^*$Corresponding author: Lijun Zhang}
}

\begin{document}

\maketitle
\thispagestyle{empty}
\pagestyle{empty}

\begin{abstract}
	In multi-vehicle cooperative driving tasks involving high-frequency continuous control, traditional state-based reward functions suffer from the issue of vanishing reward differences. This phenomenon results in a low signal-to-noise ratio (SNR) for policy gradients, significantly hindering algorithm convergence and performance improvement. To address this challenge, this paper proposes a novel Hybrid Differential Reward (HDR) mechanism. We first theoretically elucidate how the temporal quasi-steady nature of traffic states and the physical proximity of actions lead to the failure of traditional reward signals. Building on this analysis, the HDR framework innovatively integrates two complementary components: (1) a Temporal Difference Reward (TRD) based on a global potential function, which utilizes the evolutionary trend of potential energy to ensure optimal policy invariance and consistency with long-term objectives; and (2) an Action Gradient Reward (ARG), which directly measures the marginal utility of actions to provide a local guidance signal with a high SNR. Furthermore, we formulate the cooperative driving problem as a Multi-Agent Partially Observable Markov Game (POMDPG) with a time-varying agent set and provide a complete instantiation scheme for HDR within this framework. Extensive experiments conducted using both online planning (MCTS) and Multi-Agent Reinforcement Learning (QMIX, MAPPO, MADDPG) algorithms demonstrate that the HDR mechanism significantly improves convergence speed and policy stability. The results confirm that HDR guides agents to learn high-quality cooperative policies that effectively balance traffic efficiency and safety.
\end{abstract}

\section{Introduction}

As autonomous driving technology evolves from single-vehicle intelligence to collective intelligence, cooperative driving has emerged as a critical approach to mitigating traffic congestion and enhancing road safety \cite{gek2022mvsv}. Unlike the reactive decision-making of single-vehicle systems, cooperative driving requires multiple agents to engage in strategic interactions and game-theoretic reasoning to seek joint solutions that maximize collective benefits. Multi-Agent Reinforcement Learning (MARL) is widely regarded as an ideal paradigm for achieving this goal due to its capability to handle high-dimensional state spaces and complex game dynamics \cite{KherroubiAknine-208,CrosatoTian-262}.

However, despite the significant success of MARL in discrete games such as StarCraft and Go, its application to high-frequency, continuous decision-making tasks—specifically vehicle control—faces a critical yet long-overlooked challenge: the problem of vanishing reward differences \cite{LuZhao-234, TangHuang-226}.

The root of this problem lies in the inherent conflict between the temporal quasi-steady nature of the physical traffic system and the high-frequency characteristics of decision-making algorithms. Within a short decision interval (e.g., 0.1 seconds), vehicle inertia dictates that macroscopic states, such as position and velocity, undergo only negligible evolution. Traditional state-based reward functions fail to capture these subtle changes, causing the differences in reward signals generated by distinct actions (e.g., rapid acceleration versus gradual acceleration) to be overwhelmed by environmental noise. This results in an excessively low Signal-to-Noise Ratio (SNR) for policy gradients, leading to issues such as vanishing gradients or excessive variance during the early stages of training. Consequently, algorithms struggle to converge to complex cooperative policies.

Existing solutions primarily focus on reward shaping \cite{NgHarada-304} or intrinsic motivation \cite{li2021celebrating}. Although classic Potential-Based Reward Shaping (PBRS) \cite{NgHarada-304} theoretically guarantees optimal policy invariance, it fundamentally relies on state differences ($\phi(s') - \phi(s)$). When state changes are negligible, the guidance signal provided by PBRS remains equally weak, failing to fundamentally resolve the issue of low gradient SNR. In other words, existing methods address the directionality of rewards but fail to resolve the issue of reward sensitivity.

To overcome this bottleneck, this paper proposes a novel Hybrid Differential Reward (HDR) mechanism. HDR moves beyond reliance solely on state evaluation by innovatively introducing the dimension of action gradients. It integrates two complementary signals: (1) a Temporal Difference Reward (TRD) based on a potential function, which utilizes state evolution trends to ensure consistency with long-term optimization objectives and theoretical completeness; and (2) an Action Gradient Reward (ARG) based on direct action effects, which constructs a local gradient signal with high SNR and strong causality by directly measuring the marginal contribution of actions to the objective.

The main contributions of this paper are as follows:
\begin{itemize}
	\item We formally define the problem of vanishing reward differences caused by the temporal quasi-steady nature of states and the physical proximity of actions in the context of high-frequency control, revealing the fundamental reason for MARL training failures in this domain.
	\item We propose the Hybrid Differential Reward mechanism. Through theoretical proof, we demonstrate how the TRD component guarantees optimal policy invariance and how the ARG component significantly enhances gradient SNR by reshaping the local optimization landscape.
	\item We model the cooperative driving problem as a Multi-Agent Partially Observable Markov Game (POMDPG) with a time-varying agent set and provide a complete instantiation scheme for HDR within this complex task.
	\item Experiments across various paradigms, including online planning (MCTS) and offline learning (QMIX, MAPPO, MADDPG), demonstrate that HDR significantly accelerates convergence and improves policy quality, validating its effectiveness as a general reward design paradigm for high-frequency decision-making.
\end{itemize}

\section{Related Works}
\label{sec_related_works}

This section reviews the modeling paradigms for multi-vehicle cooperative driving, with a specific focus on the evolution of reward function design and the limitations encountered in continuous control tasks.

\subsection{Cooperative Driving and Game Modeling}
Cooperative driving aims to resolve complex right-of-way conflicts and optimize macroscopic traffic efficiency through active collaboration among vehicles. Early studies were predominantly based on optimal control theory, modeling the problem as centralized Model Predictive Control (MPC) \cite{GuanMeng-233,HossainLu-267}. However, these methods typically treat other vehicles as predictable dynamic obstacles, making it difficult to handle complex strategic interactions.

To characterize dynamic games among agents, Game Theory provides a more natural mathematical framework \cite{CrosatoTian-262}. Extensive research has been conducted, ranging from Stackelberg games handling asymmetric interactions \cite{HangLv-202} to Bayesian games addressing incomplete information \cite{KherroubiAknine-208}. Currently, Multi-Agent Markov Games have become the mainstream modeling paradigm due to their ability to unify the description of sequential decision-making, strategic interaction, and environmental stochasticity \cite{ChenHajidavalloo-237,ValienteToghi-277}. This paper adopts this framework, extending it to accommodate the dynamic variation in the number of agents within continuous traffic flows.

\subsection{Reward Design Dilemma and Shaping Methods}
In reinforcement learning and game solving, the reward function plays a pivotal role in guiding policy evolution. Traditional reward design typically employs a weighted sum formulation, comprehensively balancing multiple conflicting objectives such as safety, efficiency, and comfort \cite{LuZhao-228,HangHuang-210}.

However, in systems with smooth state evolution like traffic flow, researchers have found that reward functions based on sparse events (e.g., collisions) or macroscopic states (e.g., velocity) often lead to low learning efficiency \cite{Mataric-294}. Particularly under high-frequency decision-making, the state difference $\Delta s$ resulting from adjacent actions is minimal, causing the gradient $\nabla_a R$ of the reward signal $R(s, a)$ to be overwhelmed by environmental noise \cite{LiuHang-188}. This corresponds to the problem of vanishing reward differences described previously.

To address the sparse reward problem, Ng et al. proposed the renowned Potential-Based Reward Shaping (PBRS) theory \cite{NgHarada-304}, proving that an additional reward in the form of $F = \gamma \phi(s') - \phi(s)$ preserves optimal policy invariance. This theory has been widely applied to accelerate RL training \cite{Grzes-305,GuptaPacchiano-301}. Additionally, the Reward Centering technique proposed by Naik et al. \cite{NaikWan-302} reduces variance by subtracting a dynamic baseline, thereby improving learning stability.

Although the aforementioned methods are theoretically sound, they struggle to address the sensitivity issues inherent in high-frequency continuous control. PBRS fundamentally relies on the temporal difference of states; when the system is in a quasi-steady state where $\phi(s') \approx \phi(s)$, the resulting shaping signal remains weak. In other words, existing reward shaping techniques primarily focus on guiding the optimization direction correctly, but fail to adequately address the effective differentiation of subtle action variations. The HDR mechanism proposed in this paper fills this critical research gap by introducing an Action Gradient signal directly coupled with actions.

\section{Problem Formulation}
\label{sec_problem_form}

This section aims to establish a precise mathematical model for the multi-vehicle cooperative decision-making problem and to analyze the inherent structural deficiencies of traditional reward functions within this context. This analysis serves as the theoretical basis for the proposed Hybrid Differential Reward (HDR) mechanism.

\begin{figure}[h]
	\centering
	\includegraphics[width=1.0\columnwidth]{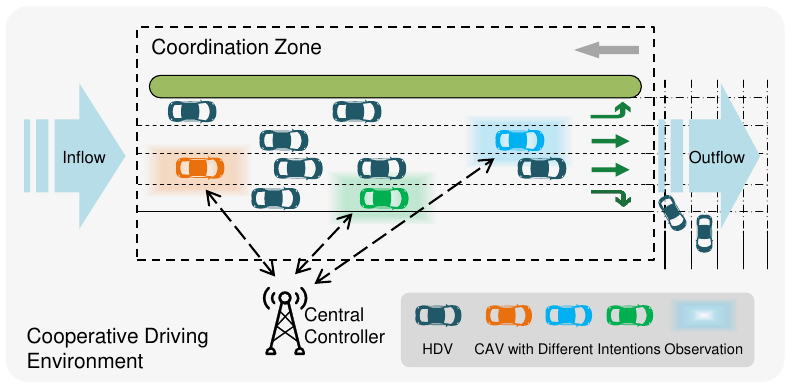}
	\caption{Schematic of the cooperative driving environment under continuous traffic flow.}
	\label{fig:basic_scene}
\end{figure}

\subsection{Scenario Description and System Characteristics}
\label{subsec_scene_desc}

This paper investigates the multi-vehicle cooperative driving problem within continuous traffic flow, as illustrated in Fig. \ref{fig:basic_scene}. The system comprises a designated coordination zone, characterized by continuous dynamic traffic with inflow from the left and outflow to the right. New vehicles continuously enter the task area while others complete their tasks and exit, resulting in a controlled agent set $\mathcal{N}_t$ that varies dynamically over time rather than remaining fixed. The road segment contains a mix of Human-Driven Vehicles (HDVs, shown in dark colors) with stochastic behaviors and controlled Connected Automated Vehicles (CAVs, shown in bright colors). As depicted, CAVs with distinct colors possess different driving intentions (e.g., the orange vehicle intends to turn left, while the green vehicle intends to go straight), leading to complex right-of-way competitions and potential conflicts within a limited spatiotemporal domain. Constrained by physical perception ranges, each CAV only possesses local observations (indicated by the halo range). A central controller is required to aggregate these local observations, infer the global state, and compute optimal joint actions in real-time to ensure global safety and efficiency under uncertainty.

\subsection{Multi-Agent Markov Game Modeling}
\label{subsec_math_mod}

Based on the aforementioned characteristics, we model the multi-vehicle cooperative driving problem as a Multi-Agent Partially Observable Markov Game (POMDPG) with a time-varying agent set \cite{kalogiannis2023towards}. The game is defined by the tuple $\mathcal{G}= \langle \{\mathcal{N}_t\}_{t\ge 0}, \mathcal{S}, \{\mathcal{A}_t\}_{t\ge 0}, P, R, \Omega, O, \gamma \rangle$, where $\mathcal{N}_t$ is the set of agents present in the coordination zone at time step $t$, a key time-varying feature. $\mathcal{S}$ denotes the global state space, containing complete information about all vehicles and the road environment. $\mathcal{A}_t$ represents the joint action space, defined as $\mathcal{A}_t = \times_{i \in \mathcal{N}_t} \mathcal{A}^{(i)}$, whose dimension varies with $\mathcal{N}_t$. $P(s'|s, \bm{a}_t)$ is the state transition function, describing vehicle dynamics, stochastic HDV behaviors, and the random processes of vehicle entry and exit. $R(s, \bm{a}_t)$ is the joint reward function, returning a global reward value based on collective performance. $O(s, i)$ is the observation function, generating a private local observation $o^{(i)} \in \Omega$ for agent $i$. $\gamma$ is the discount factor.

We adopt a centralized control paradigm based on local observations. The central decision unit outputs a joint action $\bm{a}_t \in \mathcal{A}_t$ based on the joint policy $\bm{\pi}(\bm{a}_t | \bm{o}_t)$. The discrete action space for a single agent is the Cartesian product of the longitudinal action set $\mathcal{A}^{(i)}_{\text{long}} = \{\text{AC, MT, DC}\}$ (accelerate, maintain speed, decelerate) and the lateral action set $\mathcal{A}^{(i)}_{\text{lat}} = \{\text{LC, LK, RC}\}$ (change left, keep lane, change right), i.e., $\mathcal{A}^{(i)} = \mathcal{A}^{(i)}_{\text{lat}} \times \mathcal{A}^{(i)}_{\text{long}}$.

The objective of the system is to find an optimal centralized joint policy $\bm{\pi}^{*}$ that maximizes the expected total group return:
\begin{equation}
	\bm{\pi}^{*} = \underset{\bm{\pi}}{\mathrm{argmax}} \ \mathbb{E} \left[ \sum_{t=0}^{T-1} \sum_{j \in \mathcal{N}_t} \gamma^t w_{r}^{(j)} r_t^{(j)} \Big| \bm{\pi} \right]
	\label{eq:ch_formu_obj}
\end{equation}
where $r_t^{(j)}$ is the individual reward for vehicle $j$.

\subsection{Limitations Analysis}
\label{subsec_drawback}

Although the optimization objective defined in Eq. \eqref{eq:ch_formu_obj} is macroscopically correct, applying it to high-frequency, continuous decision-making processes reveals inherent structural deficiencies that lead to the problem of vanishing reward differences. Specifically, the reward signals become too weak and ambiguous to provide effective gradients for the learning process. We analyze these limitations from two dimensions: temporal and action.

\subsubsection{Temporal Dimension}
States exhibit temporal quasi-steady characteristics. Traffic flow acts as a typical quasi-steady system where macroscopic states evolve smoothly over short intervals. Even if an agent executes an optimal action, the state at the next time step $s_{t+1}$ remains highly similar to the current state $s_t$.

\begin{figure}[htbp]
	\centering
	\includegraphics[width=1.0\columnwidth]{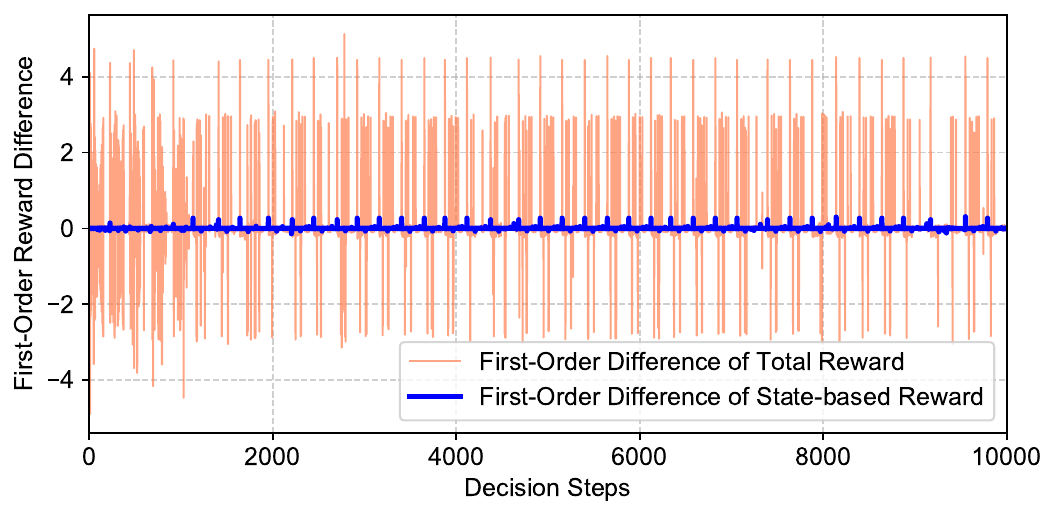}
	\caption{Schematic of state evolution for key variables (e.g., velocity, distance) under high-frequency decision-making. Despite policy differences, the state difference $\Delta s$ between $s_t$ and $s_{t+1}$ within a small time step $\Delta t$ is minimal, causing the state-based reward difference to approach zero.}
	\label{fig:state_evolution}
\end{figure}

As shown in Fig. \ref{fig:state_evolution}, this temporal quasi-steady nature causes the state-based reward difference $\Delta R_s = R(s_{t+1}) - R(s_t)$ to be extremely small, resulting in a locally flat optimization landscape. For algorithms relying on these signals for exploration and learning, such weak and sparse gradient signals are virtually indistinguishable from noise.

\begin{figure}[h]
	\centering
	\includegraphics[width=1.0\columnwidth]{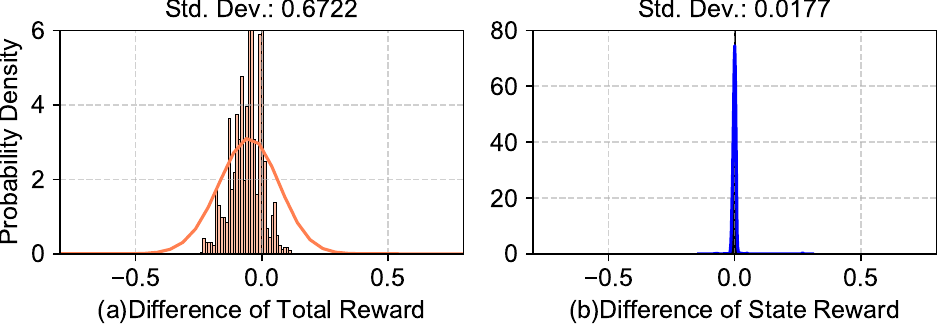}
	\caption{Comparison of distributions between the first-order difference of the state-based reward signal and the total reward signal. The reward difference $\Delta R$ is significantly smaller than the noise $\sigma$ introduced by the environment or observations, leading to a low signal-to-noise ratio in policy gradients and an inability to effectively distinguish marginal action utilities.}
	\label{fig:ch2_reward_challenge_time-2}
\end{figure}

\subsubsection{Action Dimension}
Actions exhibit physical proximity. Under high-frequency decision-making, two semantically distinct but physically adjacent discrete actions (e.g., "maintain speed" vs. "accelerate") produce negligible differences between the resulting next states $s'_{1}$ and $s'_{2}$ within a extremely short time step $\Delta t = 0.1$ s. For instance, after 0.1 s, the difference in travel distance between actions with accelerations of $0.0 \text{ m/s}^2$ and $1.0 \text{ m/s}^2$ is merely $0.005$ meters.

For a state-based reward function, this implies that the difference in calculated reward values $R(s')$ becomes infinitesimal and is easily overwhelmed by random environmental noise. This phenomenon leads to severe credit assignment problems, causing instability in the optimization process and difficulty in policy convergence. The root cause is that the reward signal evaluates only the outcome of the action—namely, the next state—rather than the quality of the action itself. Therefore, there is an urgent need for a mechanism that can explicitly measure both the temporal evolution trend of states and the marginal utility of actions, transforming a flat reward surface into a strongly guided optimization signal.

\begin{figure*}[t]
	\centering
	\includegraphics[width=1.0\textwidth]{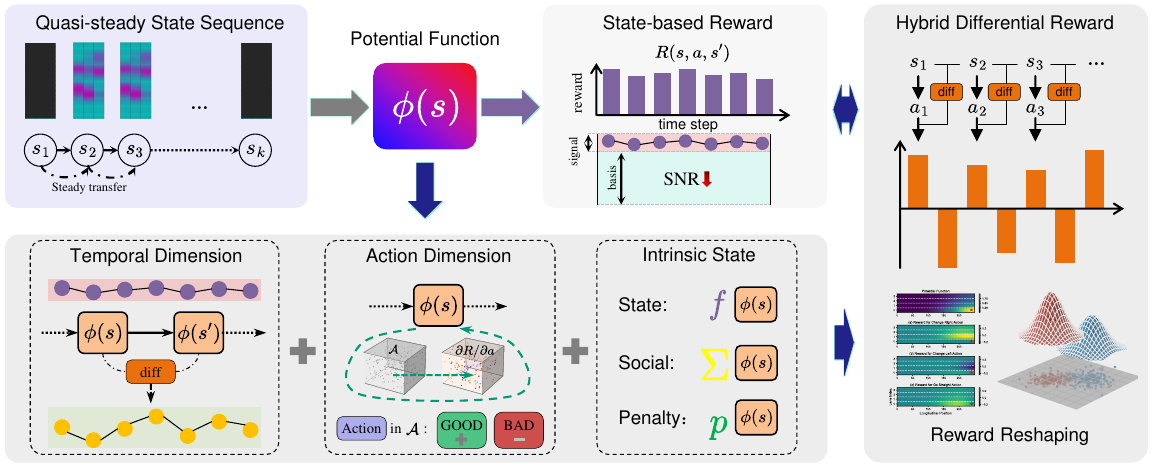}
	\caption{Overall architecture of the Hybrid Differential Reward.}
	\label{fig:hdr_schema}
\end{figure*}

\section{Methodology}
\label{sec_method}

To address the problem of vanishing reward differences described in Section \ref{subsec_drawback}, this section proposes a novel Hybrid Differential Reward (HDR) mechanism. By fusing two complementary differential signals, this mechanism ensures the correctness of long-term optimization objectives while significantly improving the signal-to-noise ratio (SNR) for distinguishing short-term actions.

\subsection{HDR Framework Construction}
\label{subsec_hdr_framework}

The core idea of the HDR mechanism is to construct an objective function $J_{\text{HDR}}$, which is a weighted combination of the Temporal Difference Reward (TRD) based on state evolution and the Action Gradient Reward (ARG) based on the marginal utility of actions.

\subsubsection{Temporal Difference Reward (TRD)}

To provide a reward signal that is oriented towards long-term goals and guarantees optimal policy invariance, we introduce the Temporal Difference Reward (TRD) based on a potential function. TRD draws upon the theory of potential-based reward shaping but applies it as a differential approximation of state value.

Based on a state potential function $\phi(s)$, we define an augmented reward function $R'$, which consists of the original environmental reward $R$ and an additional TRD signal $F_{\text{TRD}}$:
\begin{equation}
	R'(s, a, s') = R(s, a, s') + F_{\text{TRD}}(s, s'),
	\label{eq:ch2_trd_signal_full}
\end{equation}
where the TRD signal $F_{\text{TRD}}$ is defined as the discounted difference between the potential energy of the next state and the current state:
\begin{equation}
	F_{\text{TRD}}(s, s') \triangleq \gamma \phi(s') - \phi(s).
	\label{eq:ch2_trd_signal_def}
\end{equation}
Here, $\gamma$ is the discount factor. The $F_{\text{TRD}}$ signal intuitively reflects whether the system state is evolving towards a favorable direction ($F_{\text{TRD}}>0$) or an unfavorable one ($F_{\text{TRD}}<0$) after executing action $a$.

Based on this definition, we propose the following \textbf{Proposition}: For any Markov Decision Process $\mathcal{M} = (\mathcal{S}, \mathcal{A}, P, R, \gamma)$ and an augmented MDP $\mathcal{M}' = (\mathcal{S}, \mathcal{A}, P, R', \gamma)$ with the TRD signal defined in Eq. (\ref{eq:ch2_trd_signal_full}), $\mathcal{M}$ and $\mathcal{M}'$ share the same set of optimal policies. This ensures that the HDR mechanism remains consistent with the original optimization problem regarding long-term objectives.

\begin{proof}
	To concisely describe the Bellman iteration process, we introduce the Bellman optimal operator $\mathcal{T}$. For the original MDP, its application to any value function $Q$ is defined as:
	\begin{equation}
		(\mathcal{T} Q)(s, a) \triangleq \mathbb{E}_{s'} \left[ R(s, a, s') + \gamma \max_{a'} Q(s', a') \right].
	\end{equation}
	By definition, the original optimal value function $Q_{\mathcal{M}}^*$ is the unique fixed point of the operator $\mathcal{T}$, i.e., $\mathcal{T} Q_{\mathcal{M}}^* = Q_{\mathcal{M}}^*$. Similarly, the operator $\mathcal{T}'$ for the augmented MDP is defined as:
	\begin{equation}
		(\mathcal{T}' Q)(s, a) \triangleq \mathbb{E}_{s'} \left[ R'(s, a, s') + \gamma \max_{a'} Q(s', a') \right].
	\end{equation}
	
	We need to prove that the optimal value function of the augmented MDP, $Q_{\mathcal{M}'}^*$, satisfies $Q_{\mathcal{M}'}^*(s, a) = Q_{\mathcal{M}}^*(s, a) - \phi(s)$. Let the function to be verified be $\hat{Q}(s, a) \triangleq Q_{\mathcal{M}}^*(s, a) - \phi(s)$. This proposition is equivalent to verifying whether $\hat{Q}$ is a fixed point of the operator $\mathcal{T}'$.
	
	First, we introduce the original optimal state value $V_{\mathcal{M}}^*(s') \triangleq \max_{a'} Q_{\mathcal{M}}^*(s', a')$. Using the property that $\phi(s')$ is independent of actions, the optimal value of $\hat{Q}$ at the next state $s'$ can be simplified as:
	\begin{equation}
		\max_{a'} \hat{Q}(s', a')  = V_{\mathcal{M}}^*(s') - \phi(s').
		\label{eq:proof_v_star}
	\end{equation}
	
	Next, we substitute the definition of $R'$ from Eq. \eqref{eq:ch2_trd_signal_full} and Eq. \eqref{eq:proof_v_star} into the definition of operator $\mathcal{T}'$. Examining the target value $Y(s, a, s')$ inside the expectation operator, it can be decomposed into three terms:
	\begin{equation}
		R(s,a,s') + \underbrace{\gamma \phi(s') - \phi(s)}_{F_{\text{TRD}}} + \gamma \underbrace{(V_{\mathcal{M}}^*(s') - \phi(s'))}_{\max \hat{Q}(s')}
	\end{equation}
	Further simplification yields:
	\begin{equation}
		Y=R(s,a,s') + \gamma V_{\mathcal{M}}^*(s') - \phi(s).
	\end{equation}
	
	Finally, applying the expectation operator to $Y$:
	\begin{equation}
		\begin{aligned}
			(\mathcal{T}' \hat{Q})(s, a) &= \mathbb{E}_{s'} \left[ R(s,a,s') + \gamma V_{\mathcal{M}}^*(s') \right] - \phi(s) \\
			&= (\mathcal{T} Q_{\mathcal{M}}^*)(s, a) - \phi(s) \\
			&= \hat{Q}(s, a).
		\end{aligned}
	\end{equation}
	
	The derivation shows that $\hat{Q}$ is indeed a fixed point of $\mathcal{T}'$. Since $\phi(s)$ is independent of the current action $a$, the augmented policy $\pi'^*(s) = \arg\max_a \hat{Q}(s,a)$ shares the same maximizing parameters as the original policy $\pi^*(s)$. The proposition is proved.
\end{proof}

\subsubsection{Action Gradient Reward (ARG)}
\label{sec:action_gradient_reward}

Although TRD addresses the correctness of long-term objectives, it relies on macroscopic state changes $s'$, which limits its effectiveness in resolving the vanishing reward difference problem caused by the physical proximity of actions. Therefore, we propose the Action Gradient Reward (ARG), which aims to directly evaluate the marginal utility of actions.

The core of ARG is to assess the additional return gain an action $a$ provides compared to an average or baseline action $a_{\text{base}}$ in the current state $s$. This signal $r_{\text{ARG}}$ is constructed as an instantaneous reward strongly correlated with the current action $a$ and possessing high causality. It serves to reshape the local geometry of the policy gradient surface:
\begin{equation}
	r_{\text{ARG}}(s, a) \propto \nabla_a Q_{\text{local}}(s, a)
	\label{eq:arg_def}
\end{equation}
where $Q_{\text{local}}$ is a Q-function measuring short-term action utility. In practice, the ARG signal $r_{\text{ARG}}$ is designed to directly measure the immediate impact of action $a$ on the potential function $\phi(s)$ or its reduction of the instantaneous cost function. The introduction of ARG ensures that even when the difference between states $s_{t}$ and $s_{t+1}$ is minimal, small but goal-oriented actions $a$ receive significant rewards, thereby significantly improving the SNR of policy gradients.

\subsubsection{Hybrid Optimization Objective}

A standard MDP optimization problem aims to maximize the expected cumulative return, which can be expressed as $J_R(\theta) = \mathbb{E}_{\tau \sim \pi_\theta} \left[ \sum_{k=0}^{\infty} \gamma^k R(s_k, a_k) \right]$. We decompose and reconstruct this objective to reflect the two differential concepts.

First, to explicitly incorporate the TRD signal, we define the first objective function: the State Potential Elevation Objective $J_{\text{TRD}}(\theta)$. According to the TRD proposition, the goal of the signal $\gamma\phi(s') - \phi(s)$ is to guide the agent towards states with higher potential energy. Therefore, maximizing the expectation of the TRD reward is equivalent to maximizing the total potential energy gain brought by the policy over the entire trajectory:
\begin{equation}
	J_{\text{TRD}}(\theta) \triangleq \mathbb{E}_{\tau \sim \pi_\theta} \left[ \sum_{k=0}^{\infty} \gamma^k (\gamma\phi(s_{k+1}) - \phi(s_k)) \right]
\end{equation}
This objective explicitly sets the pursuit of favorable state evolution trends as the optimization direction.

To explicitly incorporate the ARG signal, we define the second objective function: the Action-Reward Alignment Objective $J_{\text{ARG}}(\theta)$. Its goal is to maximize the consistency between the selected action and the direction of the reward gradient. A direct definition is to maximize the expectation of the reward function, as its gradient is directly related to the ARG signal:
\begin{equation}
	J_{\text{ARG}}(\theta) = \mathbb{E}_{s \sim \rho^{\pi_\theta}, a \sim \pi_\theta(a|s)} \left[ R(s, a) \right].
\end{equation}

Combining these two objectives linearly, we form the foundational form of the objective function under the hybrid differential reward framework:
\begin{equation}
	\label{eq:hdr_objective}
	J_{\text{HDR}}(\theta) \triangleq (1-\alpha) J_{\text{TRD}}(\theta) + \alpha J_{\text{ARG}}(\theta),
\end{equation}
where $\alpha \in [0, 1]$ is a hyperparameter used to balance the importance of long-term returns and immediate action guidance.

Having determined the hybrid optimization objective, the core of any gradient-ascent-based algorithm is to calculate the policy gradient of the objective function. Following Eq. \eqref{eq:hdr_objective}, the policy gradient for HDR can be expressed as:
\begin{equation}
	\nabla_\theta J_\text{HDR}(\theta) = (1-\alpha) \nabla_\theta J_{\text{TRD}}(\theta) + \alpha \nabla_\theta J_{\text{ARG}}(\theta).
\end{equation}
We now analyze the specific forms of these two gradient terms.

For the State Potential Elevation Objective $J_{\text{TRD}}(\theta)$, the single-step reward is $r_{\text{TRD}}(s,a,s') = \gamma\phi(s') - \phi(s)$. According to the Policy Gradient Theorem, its gradient can be written as:
$\nabla_\theta J_{\text{TRD}}(\theta) = \mathbb{E}_{\tau \sim \pi_\theta} \left[ \sum_{k=0}^{\infty} \nabla_\theta \log \pi_\theta(a_k|s_k) A_{\text{TRD}}^{\pi_\theta}(s_k, a_k) \right]$,
where $A_{\text{TRD}}^{\pi_\theta}(s, a)$ is the advantage function corresponding to the TRD reward under policy $\pi_\theta$.

For the Action-Reward Alignment Objective $J_{\text{ARG}}(\theta)$, according to the Deterministic Policy Gradient Theorem \cite{silver2014deterministic}, its gradient for a deterministic policy $\pi_\theta(s)$ is:
$\nabla_\theta J_{\text{ARG}}(\theta) = \mathbb{E}_{s \sim \rho^{\pi_\theta}} \left[ \nabla_\theta \pi_\theta(s) \nabla_a R(s, a) \Big|_{a=\pi_\theta(s)} \right]$.

Combining these yields the general policy gradient under the HDR framework. The HDR framework is algorithm-agnostic; in principle, any algorithm capable of performing policy optimization can apply this framework. For example, in Actor-Critic methods, the optimization of the Actor can be achieved by superimposing these two gradient terms. In deterministic policy methods like DDPG and TD3, the loss function for the Actor can be designed as:
$\mathcal{L}_{actor}(\theta) = -\mathbb{E}_s \left[ (1-\alpha)A_{\text{TRD}}\left(s, \pi_\theta(s)\right) + \alpha R\left(s, \pi_\theta(s)\right) \right]$.
Taking the negative gradient of this loss function yields the update direction consistent with the HDR objective.

\textbf{Assumptions for HDR Objective Convergence:} The following three conditions are assumed:
1) Policy Smoothness: The policy $\pi_\theta(s)$ is continuously differentiable with respect to its parameters $\theta$. For deterministic policies, the output $\pi_\theta(s)$ and the Jacobian matrix $\nabla_\theta \pi_\theta(s)$ are $L$-Lipschitz continuous with respect to $\theta$.
2) Potential Function and TRD Objective Smoothness: The potential function $\phi(s)$ is bounded and differentiable. Consequently, the defined State Potential Elevation Objective $J_{\text{TRD}}(\theta)$ is continuously differentiable, and its gradient $\nabla_\theta J_{\text{TRD}}(\theta)$ is $L_{\text{TRD}}$-Lipschitz continuous.
3) Reward Function and ARG Objective Smoothness: The reward function $R(s, a)$ is continuously differentiable with respect to action $a$. Its gradient, i.e., the ARG signal $R'_a(s,a) = \nabla_a R(s, a)$, has a bounded norm ($\norm{R'_a(s,a)} \le C_R < \infty$) and is $L_a$-Lipschitz continuous with respect to action $a$.

Under these assumptions, the gradient of the Hybrid Differential Reward objective function $J_{\text{HDR}}(\theta)$ is $L$-Lipschitz continuous. Therefore, any algorithm performing gradient ascent on $J_{\text{HDR}}(\theta)$, with an appropriate learning rate, will generate a parameter sequence $\{\theta_k\}$ that converges to a stationary point satisfying $\nabla_\theta J_{\text{HDR}}(\theta^*) = 0$.

\begin{proof}
	The goal is to prove that the gradient $\nabla_\theta J_{\text{HDR}}(\theta)$ is $L$-Lipschitz continuous. If proven, the convergence of the gradient ascent algorithm is guaranteed by standard optimization theory. The proof proceeds in three steps:
	
	Step 1: Analyze the properties of $\nabla_\theta J_{\text{TRD}}(\theta)$.
	According to the assumptions, $\nabla_\theta J_{\text{TRD}}(\theta)$ is already $L_{\text{TRD}}$-Lipschitz continuous.
	
	Step 2: Prove the Lipschitz continuity of $\nabla_\theta J_{\text{ARG}}(\theta)$.
	Define the gradient coupling term $\mathcal{O}(\theta, s)$ as the product of the policy Jacobian and the action gradient:
	\begin{equation}
		\mathcal{O}(\theta, s) \triangleq \underbrace{(\nabla_\theta \pi_\theta(s))^T}_{A(\theta)} \underbrace{\nabla_a R(s, \pi_\theta(s))}_{B(\theta)}.
	\end{equation}
	The objective gradient can be simplified as $\nabla_\theta J_{\text{ARG}}(\theta) = \mathbb{E}_{s} [\mathcal{O}(\theta, s)]$. Consider the norm of the gradient difference and apply Jensen's inequality to extract the expectation:
	\begin{equation}
		\begin{aligned}
			&\norm{\nabla_\theta J_{\text{ARG}}(\theta_1) - \nabla_\theta J_{\text{ARG}}(\theta_2)} \\
			\le &\mathbb{E}_{s} \norm{\mathcal{O}(\theta_1, s) - \mathcal{O}(\theta_2, s)}.
		\end{aligned}
	\end{equation}
	
	For the internal term $\mathcal{O}$, using the triangle inequality and decomposition identity:
	\begin{equation}
		\begin{aligned}
			& \norm{\mathcal{O}(\theta_1) - \mathcal{O}(\theta_2)} \\
			= & \norm{A(\theta_1)B(\theta_1) - A(\theta_2)B(\theta_2)} \\
			\le & \norm{A(\theta_1)} \norm{B(\theta_1) - B(\theta_2)} +\\
			& \norm{A(\theta_1) - A(\theta_2)} \norm{B(\theta_2)}.
		\end{aligned}
	\end{equation}
	
	Based on the boundedness (bounds $C_A, C_B$) and Lipschitz continuity (constants $L_A, L_B$) from the assumptions, we can directly derive the upper bounds for each term. Specifically for the composite function $B(\theta)$, using the chain rule property, we have $\norm{B(\theta_1) - B(\theta_2)} \le L_a L_\pi \norm{\theta_1 - \theta_2}$. Substituting into the above inequality:
	\begin{equation}
		\begin{aligned}
			&\norm{\mathcal{O}(\theta_1) - \mathcal{O}(\theta_2)} \\
			\le & \left( C_A (L_a L_\pi) + L_A C_B \right) \norm{\theta_1 - \theta_2} \\
			\triangleq & L_{\text{ARG}} \norm{\theta_1 - \theta_2}.
		\end{aligned}
	\end{equation}
	This proves that $\nabla_\theta J_{\text{ARG}}$ is $L_{\text{ARG}}$-Lipschitz continuous.
	
	Step 3: Prove the Lipschitz continuity of $\nabla_\theta J_{\text{HDR}}(\theta)$. Analyzing the Lipschitz property of the entire HDR gradient:
	\begin{equation}
		\begin{aligned}
			& \norm{\nabla_\theta J_\text{HDR}(\theta_1) - \nabla_\theta J_\text{HDR}(\theta_2)} \\
			\le & (1-\alpha)\norm{\nabla_\theta J_{\text{TRD}}(\theta_1) - \nabla_\theta J_{\text{TRD}}(\theta_2)} + \\
			&\alpha\norm{\nabla_\theta J_{\text{ARG}}(\theta_1) - \nabla_\theta J_{\text{ARG}}(\theta_2)} \\
			\le & (1-\alpha) L_{\text{TRD}} \norm{\theta_1 - \theta_2} + \alpha L_{\text{ARG}} \norm{\theta_1 - \theta_2} \\
			= & \left( (1-\alpha) L_{\text{TRD}} + \alpha L_{\text{ARG}} \right) \norm{\theta_1 - \theta_2}.
		\end{aligned}
	\end{equation}
	
	Let $L_\text{HDR} = (1-\alpha) L_{\text{TRD}} + \alpha L_{\text{ARG}}$. This proves that $\nabla_\theta J_\text{HDR}(\theta)$ is $L_\text{HDR}$-Lipschitz continuous.
	
	According to standard results in optimization theory, for an objective function $J(\theta)$, if its gradient $\nabla_\theta J(\theta)$ is $L$-Lipschitz continuous, then for a gradient ascent algorithm $\theta_{k+1} = \theta_k + \eta \nabla_\theta J(\theta_k)$, selecting a learning rate $\eta \in (0, 2/L)$ guarantees $\lim_{k \to \infty} \norm{\nabla_\theta J(\theta_k)} = 0$, meaning the sequence converges to a stationary point. Since we have proven that $\nabla_\theta J_\text{HDR}(\theta)$ is $L_\text{HDR}$-Lipschitz continuous, any algorithm based on gradient ascent optimizing $J_\text{HDR}(\theta)$ is guaranteed to converge under the appropriate learning rate condition. The proof is complete.
\end{proof}

\subsection{HDR Instantiation}
\label{subsec_hdr_case}

We specifically instantiate the HDR theoretical framework for the multi-vehicle cooperative driving task to obtain a computable reward function $r_{\text{HDR}}^{(i)}$.

\subsubsection{Design of Driving Potential Function $\phi(s)$}

The implementation of the TRD mechanism requires a signal that effectively guides the policy towards the desired state evolution. The foundation is to define a potential function $\phi(s)$ that quantifies state superiority or value. For vehicle $i$, the value of its state is primarily determined by two dimensions: longitudinal (efficiency in the driving direction) and lateral (ability to maintain the target lane).

Based on these two key dimensions, we define the driving potential function $\phi(s^{(i)})$ for vehicle $i$ as follows, representing the positional superiority of the vehicle in state $s^{(i)}$:
\begin{equation}
	\label{eq:potential_function}
	\phi(s^{(i)}) \triangleq \frac{ \exp{\left(-\frac{\left(x_{\text{goal}} - x^{(i)}\right)^2}{2\sigma^2}\right)} }{ \zeta \left|y_{\text{targ}}^{(i)} - y^{(i)}\right| + 1 }.
\end{equation}
The numerator in Eq. \eqref{eq:potential_function} is a Gaussian-like function that encourages the vehicle to travel towards the distant longitudinal target position $x_{\text{goal}}$. The closer the vehicle is to the target, the larger this value becomes. Parameter $\sigma$ controls the longitudinal decay scale of the potential field. The denominator is a linear penalty term encouraging the vehicle to stay in the target lane $y_{\text{targ}}^{(i)}$. The greater the deviation of the vehicle's lateral position $y^{(i)}$ from the target lane, the lower the potential function value. Parameter $\zeta$ controls the intensity of the lateral deviation penalty.

\subsubsection{Derivation of Computable TRD Signal}

Based on the potential function above, we design the TRD reward as the total derivative of the potential function $\phi$ with respect to time. Physically, this is equivalent to the power gained by the vehicle moving within the potential field:
\begin{equation}
	\label{eq:potential_gradient_reward}
	r_{\text{TRD}}^{(i)} \triangleq \frac{\mathrm{d}\phi(s^{(i)}(t))}{\mathrm{d}t} = \nabla \phi(s^{(i)}) \cdot {\bm{v}^{(i)}},
\end{equation}
where $\bm{v}^{(i)} = (v_{x}^{(i)}, v_{y}^{(i)})$ is the velocity vector of vehicle $i$.

From a finite difference perspective, $r_{\text{TRD}}^{(i)} \approx \frac{\phi(s_{t+1}) - \phi(s_t)}{\Delta t}$. It directly measures the rate of potential energy gain resulting from state changes. It serves as a high-quality reward shaping signal that preserves optimal policy invariance, embodying the TRD concept of focusing on state evolution trends.

To calculate $r_{\text{TRD}}^{(i)}$ in simulation or implementation, the theoretical dot product must be converted into specific algebraic expressions. First, combining Eq. \eqref{eq:potential_function}, we calculate the partial derivative of $\phi(s^{(i)})$ with respect to position $x$:
\begin{equation}
	\frac{\partial \phi}{\partial x} = \phi(s^{(i)}) \cdot \frac{x_{\text{goal}} - x}{\sigma^2};
\end{equation}
Next, we calculate the partial derivative of $\phi(s^{(i)})$ with respect to position $y$ under discrete coordinates:
\begin{equation}
	\frac{\partial \phi}{\partial y} = \phi(s_i) \cdot \frac{\zeta \cdot \text{sign}(\Delta y^{(i)})}{\zeta |\Delta y^{(i)}| + 1},
\end{equation}
where $\Delta y^{(i)} = y - y_{\text{targ}}^{(i)}$.

Substituting the two terms $\partial \phi / \partial x$ and $\partial \phi / \partial y$ into Eq. \eqref{eq:potential_gradient_reward} yields the computable discrete form.

\subsubsection{Discrete Action Gradient}

According to the theoretical definition in Section \ref{sec:action_gradient_reward}, the ARG signal $R'_a(s, a) = \nabla_a R(s, a)$ aims to provide gradient guidance along the direction of steepest ascent of the reward function. However, the action space $\mathcal{A}$ in this study is discrete, making direct computation of continuous derivatives with respect to actions infeasible. To retain the theoretical properties of ARG in a discrete decision space, we propose an approximate implementation method based on gradient direction alignment.

From a physical perspective, the gradient of the efficiency objective $R_{\text{flow}}$ with respect to longitudinal acceleration $a$, $\frac{\partial R}{\partial a}$, typically exhibits a positive correlation (i.e., positive acceleration improves system efficiency, provided it is safe and within speed limits). Therefore, although precise gradient values cannot be calculated, we can capture the \textbf{direction information} of the gradient. We instantiate ARG as an indicator function to evaluate whether the currently selected discrete action projects into the ascending half-space of the theoretical gradient.
Specifically, if the agent selects an acceleration action ($a \in \mathcal{A}_{\text{AC}}$) or maintains speed while in a high-speed state ($a \in \mathcal{A}_{\text{MT}}$), it is considered consistent with the ideal gradient direction (i.e., $\nabla_a R \cdot \vec{a} > 0$). Based on this logic, we design the ARG signal as the following gradient sign indicator:
\begin{equation}
	r_{\text{ARG}}^{(i)} = 
	\begin{cases}
		1 & \text{if } a^{(i)} \in \mathcal{A}_{\text{AC}} \lor (a^{(i)} \in \mathcal{A}_{\text{MT}} \land v^{(i)} \ge v_{\text{thres}}), \\
		0 & \text{otherwise}.
	\end{cases}
	\label{eq:arg_discrete_impl}
\end{equation}
where $\mathcal{A}_{\text{AC}}$ and $\mathcal{A}_{\text{MT}}$ represent the sets of acceleration and constant speed actions, respectively, and $v_{\text{thres}}$ is a preset threshold for high-efficiency speed (set to $28 \text{m/s}$ in this experiment).

The core advantages of this design are:
1) Preservation of Causality: It discards the magnitude details of the gradient but retains the crucial "directional sign," directly informing the agent which action categories align with the local optimal gradient.
2) Robustness: Similar to the Sign-SGD method in optimization theory, guiding solely with gradient signs often provides better robustness than using high-variance precise gradient values, especially in noise-filled multi-agent interaction environments.
3) Computational Efficiency: As an analytical binary reward, its computational overhead is negligible.

\subsubsection{Complete HDR Reward Function}

Combining the TRD reward and ARG reward linearly yields the HDR reward:
\begin{equation}
	r_{\text{HDR}}^{(i)} = w_{\text{TRD}} r_{\text{TRD}}^{(i)} + w_{\text{ARG}} r_{\text{ARG}}^{(i)},
\end{equation}
where the two weight coefficients satisfy $w_{\text{TRD}} + w_{\text{TRD}} = 1$. Additionally, to complement the HDR reward, we introduce three auxiliary rewards for optimizing macroscopic traffic objectives:

1) Traffic Efficiency Reward $r_{\text{flow}}$: Evaluates the macroscopic operating efficiency of the entire traffic system, defined as the normalized average speed of all vehicles in the scene:
$	r_{\text{flow}} =  \frac{1}{|\mathcal{N}|} \sum_{j \in \mathcal{N}}\frac{v^{(j)}}{v_{\text{max}}}.$
Although the HDR reward already provides an optimization target for the overall efficiency of CAVs, CAV behavior also affects HDV driving performance. Therefore, this term is introduced as a supplement for overall traffic efficiency.

2) Safety Penalty $r_{\text{safe}}$: Penalizes dangerous driving behaviors with collision risks. Unlike sparse collision indicator functions, we design a dense penalty signal based on Time-to-Collision (TTC) to provide more forward-looking safety guidance. For each vehicle $j$, its TTC with the preceding vehicle is denoted as $\text{TTC}^{(j)}$. When $v^{(j)} > v_{h}^{(j)}$ and $H^{(j)} > 0$, its value is $ \frac{H^{(j)}}{v^{(j)} - v_{h}^{(j)}}$; otherwise, it is $\infty$.
$H^{(j)}$ is the distance between vehicle $j$ and the preceding vehicle, and $v^{(j)}$ and $v_{h}^{(j)}$ are their respective velocities. We set a safety penalty threshold $\text{TTC}_{\text{crit}}$. A safety penalty is triggered only when the vehicle's TTC falls below this threshold; higher risk results in a more severe penalty. The safety penalty for a single vehicle $r_{\text{safe}}^{(j)}$ is defined as $-1 + \exp \left( \frac{1}{\text{TTC}_{\text{crit}}} - \frac{1}{\text{TTC}^{(j)}} \right)$ if $0 < \text{TTC}^{(j)} < \text{TTC}_{\text{crit}}$; it takes the value 1 if a collision occurs, and 0 otherwise.
Based on this, the total safety penalty is the sum of penalties for all vehicles in the coordination zone, i.e., $r_{\text{safe}} = \sum_{j \in \mathcal{N}} r_{\text{safe}}^{(j)}$.

3) Frequent Lane Change Penalty $r_{\text{freq}}$: Penalizes excessively frequent lane-changing behaviors to encourage smooth driving. Unlike hard constraints using fixed time windows, we design a continuous penalty function that decays over time: $r_{\text{freq}}^{(j)} = - e^{-\lambda_{\text{LC}} \cdot t_{\text{LC${}^-$}}^{(j)}}$,
where $t_{\text{LC${}^-$}}^{(j)}$ is the time elapsed since agent $j$'s last successful lane change, and $\lambda_{\text{LC}}$ is the coefficient controlling the decay rate of the penalty. A maximum penalty is applied immediately after a lane change, and the penalty term rapidly approaches 0 through exponential decay over time.

Linearly weighting the differential reward term $r_{\text{HDR}}^{(i)}$ with the three metrics measuring global traffic efficiency, safety, and smoothness yields the complete global reward value $r_t$ for cooperative driving at time $t$:
\begin{equation}
	\begin{split}
		r_t = & \frac{1}{|\mathcal{N}_{\text{CAV}}|} \sum_{i \in \mathcal{N}_{\text{CAV}}} \left(w_{\text{HDR}} r_{\text{HDR}}^{(i)} + w_{\text{freq}} r_{\text{freq}}^{(i)}\right) \\
		& + w_{\text{flow}} r_{\text{flow}} + w_{\text{safe}} r_{\text{safe}},
	\end{split}
	\label{eq:ch_formu_reward_final}
\end{equation}
where $w_{\text{HDR}}$, $w_{\text{flow}}$, $w_{\text{safe}}$, and $w_{\text{freq}}$ are the weight coefficients for each term.

\subsubsection{Visualization Analysis of Reward Surface}

\begin{figure}[b]
	\includegraphics[width=1.0\columnwidth]{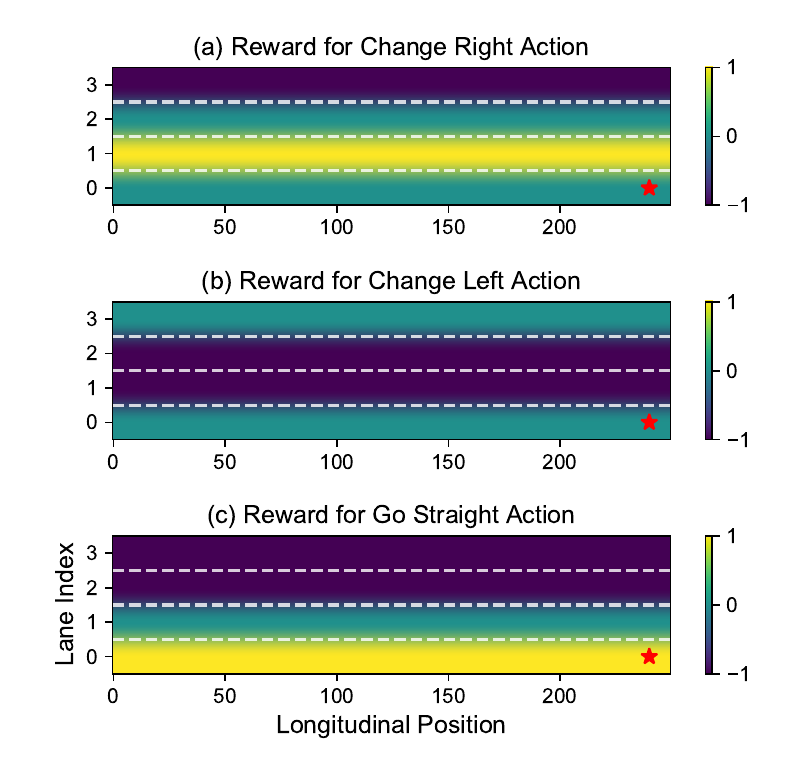} 
	\caption{State-based reward function}
	\label{fig:reward_visualization-a}
\end{figure}

To intuitively demonstrate the reward mechanism of the HDR function, Fig. \ref{fig:reward_visualization-a} compares it visually with the state-based reward function from literature \cite{grl-ini}. The scenario is set as a one-way four-lane highway, where the agent's goal is to reach the rightmost lane (Lane 0, indicated by the red star) at the end of the segment. Each subplot is a reward heatmap, with the horizontal axis representing longitudinal position and the vertical axis representing lane index. The color represents the instantaneous reward value obtained by executing a specific action at the corresponding location.

\begin{figure}[h]
	\includegraphics[width=1.0\columnwidth]{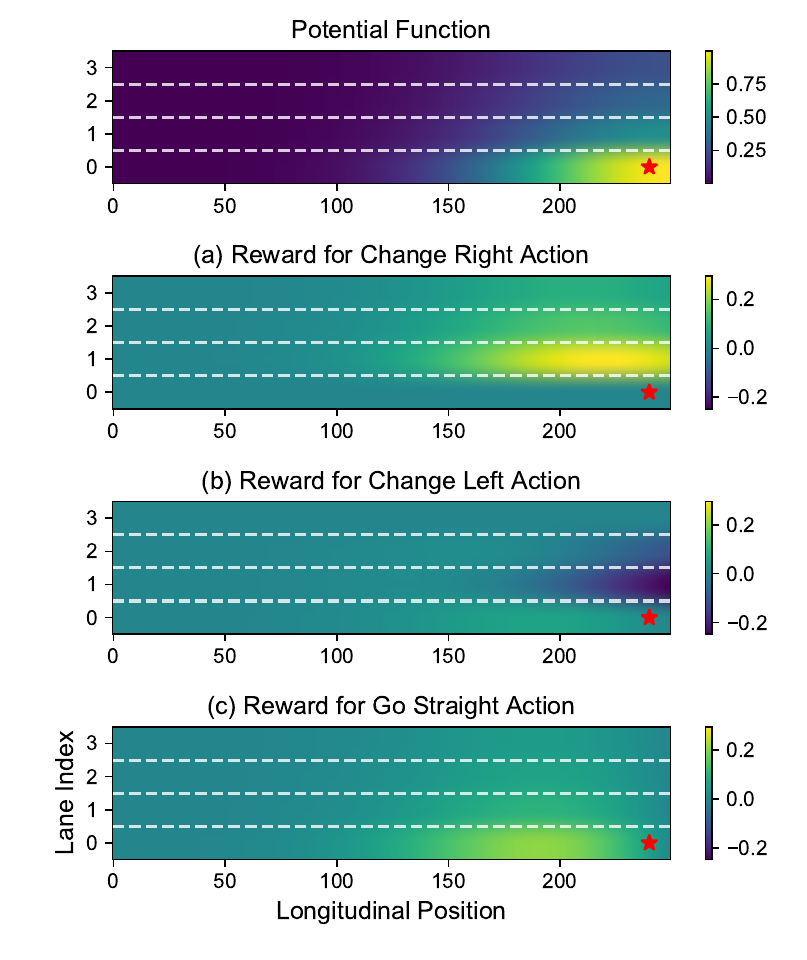} 
	\caption{HDR reward function}
	\label{fig:reward_visualization-b}
\end{figure}

In Fig. \ref{fig:reward_visualization-a}, the state-based reward function assigns values based solely on the vehicle's distance to the target after action execution. Its deficiency lies in the fact that for any fixed location, executing three different actions—"change right," "change left," or "go straight"—yields identical reward values provided the resulting position is the same. For instance, in the case of \cite{grl-ini}, the reward function penalizes vehicles located in the two rightmost lanes. Consequently, when a vehicle is in Lane 3, both "go straight" and "change right" result in penalization based on the final position, failing to provide the critical action guidance that "change right" should be executed. This lack of action differentiation in the reward signal is the root cause of low learning efficiency.

In contrast, Fig. \ref{fig:reward_visualization-b} displays the reward surface generated by the HDR reward function (primarily $r_{\text{TRD}}^{(i)}$) proposed in this paper. The potential function plot at the top shows the potential field designed by Eq. \eqref{eq:potential_function}, with energy peaking at the target point. Based on this potential field, HDR generates completely distinct reward signals for different actions: (1) The "change right" action, moving the vehicle towards areas of higher potential, receives high positive rewards in most regions; (2) Conversely, the "change left" action, moving away from high potential areas, receives generally low rewards; (3) The "go straight" action provides moderate rewards depending on the current lane. Furthermore, it can be observed that the HDR reward function proposed in this chapter does not provide explicit lateral guidance at the beginning of the road, encouraging vehicles to explore a more diverse range of actions.

In summary, the HDR framework transforms a static, action-independent state evaluation function into a dynamic gradient guidance field tightly coupled with actions. It provides decision signals on what action to execute at every point in space, thereby resolving the deficiencies of traditional methods.

\section{Experimental Verification and Analysis}
\label{sec_case_study}

This section aims to empirically analyze the proposed HDR mechanism through rigorous simulation experiments. The experiments first validate the effectiveness of the underlying world model, followed by a comparative performance analysis of HDR against existing reward designs across different decision-making algorithms (MCTS and MARL).

\subsection{Experimental Scenario and Parameter Configuration}
\label{subsec_case_setting}

The simulation environment for this study is built upon the Flow computational framework and the SUMO simulation platform. The main scenario consists of a $250 \text{m}$ long one-way four-lane continuous traffic flow. The traffic is composed of a mix of Connected Automated Vehicles (CAVs) and Human-Driven Vehicles (HDVs). During the stochastic generation of all vehicles, the intentions of turning left, going straight, and turning right are each assigned with a probability of 1/3. Regarding the functionality of the four lanes: the rightmost Lane 0 is for turning right or going straight; the middle Lanes 1 and 2 are for going straight; and the leftmost Lane 3 is for turning left or going straight. HDV behaviors are driven by the improved Krauss model (longitudinal car-following) and the LC2013 model (lateral lane-changing). The decision frequency for CAVs is set to $10 \text{ Hz}$. Key parameters for the unified simulation environment are listed in Table \ref{tab:unified_sim_params}.

\begin{table}[h]
	\centering
	\caption{Detailed Parameter Settings for Simulation Experiments}
	\label{tab:unified_sim_params}
	\begin{tabular}{ll}
		\toprule
		\multicolumn{1}{c}{\textbf{Symbol}}& \multicolumn{1}{c}{\textbf{Value}} \\ \midrule
		$v_{\text{HDV}, 0}$, $v_{\text{CAV}, 0}$& 8-12 m/s, Uniform Distribution \\
		$v_{\text{HDV},\text{max}}$, $v_{\text{CAV},\text{max}}$  & 30 m/s, 30 m/s\\
		$b$, $t_k$, $\varepsilon$& 9 m/s\textsuperscript{2}, 1.1 s, 0.5 \\
		$a_{\text{CAV}, \text{max}}$& 3.5 m/s\textsuperscript{2}\\
		$v_{\text{thres}}$& 28 m/s \\
		$\gamma$&0.996\\
		$\sigma$, $\zeta$& 60.0, 1.0 \\
		$\text{TTC}_{\text{crit}}$& 3s\\
		$\lambda_{\text{LC}}$& 0.75 \\
		$w_{\text{TRD}}$, $w_{\text{ARG}}$& 0.9, 0.1\\
		$w_{\text{HDR}}$, $w_{\text{flow}}$, $w_{\text{safe}}$, $w_{\text{freq}}$& 10.0, 1.0, 2.0, 0.9 \\
		\bottomrule
	\end{tabular}
\end{table}

\subsection{Comparative Methods and Evaluation Metrics}
\label{subsec_compa_meth}

\subsubsection{Comparative Methods}
To comprehensively evaluate the universality of the HDR mechanism, we compare it with two representative reward designs:
\begin{itemize}
	\item \textbf{General Reward (GNR)}: A traditional reward function designed based on macroscopic metrics of the current state (e.g., speed, safety).
	\item \textbf{Centering Reward (CTR)}: An advanced method that reduces learning process variance by subtracting a dynamically estimated reward baseline from the immediate reward.
	\item \textbf{Centered HDR (CTH)}: Applies the centering method to the HDR signal.
\end{itemize}
We apply the HDR, GNR, and CTR/CTH mechanisms to two mainstream decision-making paradigms: the online planning-based MCTS algorithm and policy learning-based MARL algorithms (QMIX, MAPPO, and MADDPG).

\subsubsection{Evaluation Metrics}
This paper adopts a unified evaluation metric system to assess algorithms across 5 categories comprising 8 objective indicators:
\textbf{Efficiency Metrics:} Instant Traffic Flow Rate ($\text{Inst.Flow}$) and Average Velocity ($\text{Avg.Velo.}$).
\textbf{Safety Metrics:} Average Minimum Time-to-Collision ($\text{TTC}$) and Average Number of Collisions ($\text{Coll.}$).
\textbf{Task Metrics:} Task Success Rate ($\text{Succ.Rate}$).
\textbf{Stability/Comprehensive Metrics:} Average Absolute Jerk ($\text{Jerk}$), Average Lane Change Interval ($\text{LC.Intvl.}$), and Average Traffic Score ($\text{ATS}$).

\subsection{Experimental Results and Analysis}
\label{subsec_exp_result}

\subsubsection{Validation of Effectiveness on Policy Learning (MARL)}

The experiments in this subsection aim to verify whether the optimization objective provided by HDR can enhance learning efficiency and final policy performance. MARL algorithms need to learn a policy network capable of generalizing to all states through extensive trial-and-error exploration from reward signals. Therefore, the core of this experiment is to examine which reward function can provide MARL algorithms with more stable and efficient gradients, assisting them in converging faster to higher-performance policies.

We tested three mainstream multi-agent reinforcement learning algorithms: QMIX, MAPPO, and MADDPG. All algorithms utilized unified hyperparameter settings.

\begin{figure*}[p]
	\centering
	\includegraphics[width=1.0\linewidth]{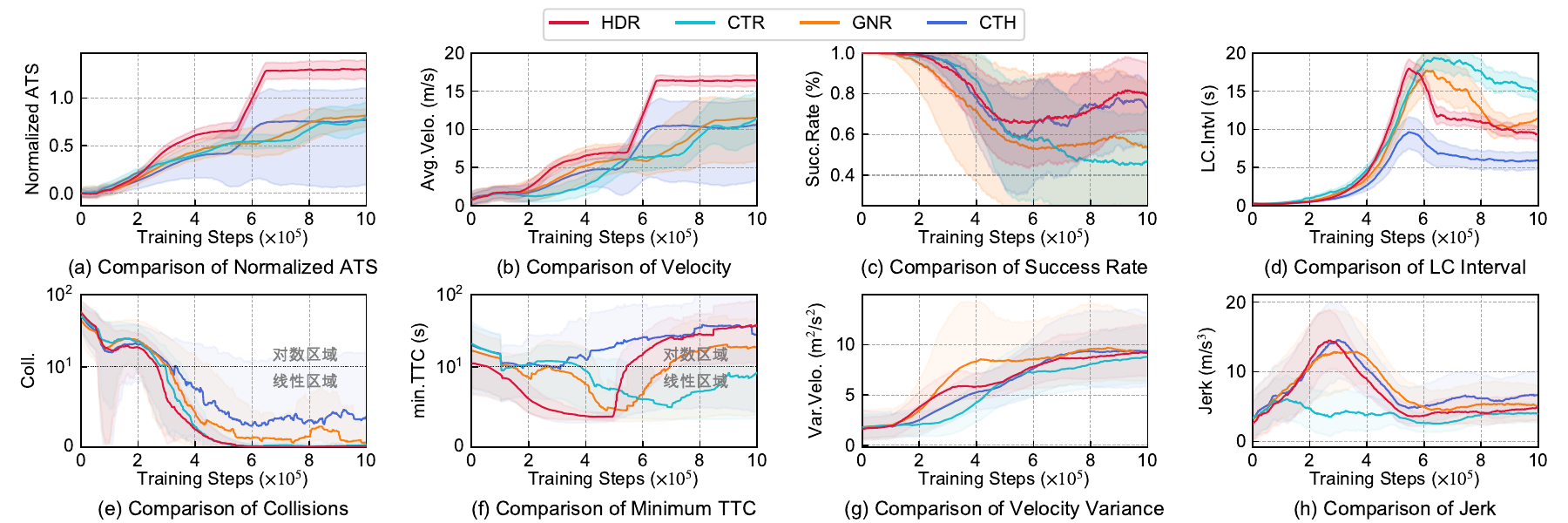}
	\caption{Performance of the QMIX algorithm guided by different reward functions.}
	\label{fig_marl_rew_comp_QMIX}
\end{figure*}

\begin{figure*}[p]
	\centering
	\includegraphics[width=1.0\linewidth]{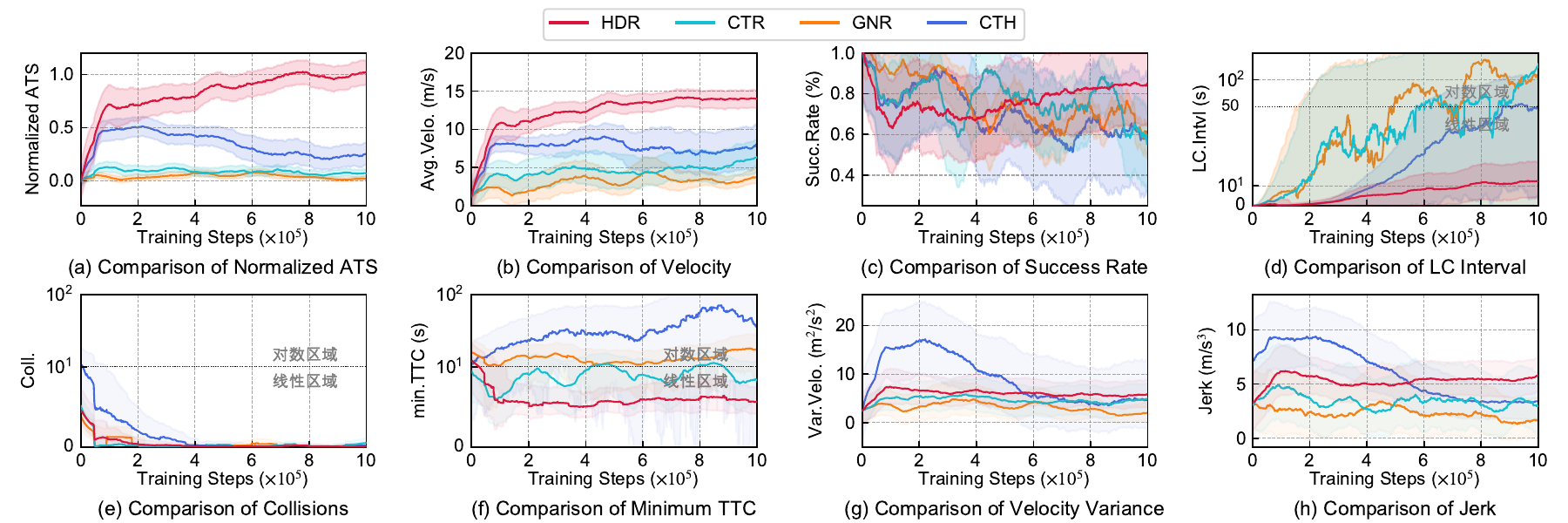}
	\caption{Performance of the MAPPO algorithm guided by different reward functions.}
	\label{fig_marl_rew_comp_MAPPO}
\end{figure*}

\begin{figure*}[p]
	\centering
	\includegraphics[width=1.0\linewidth]{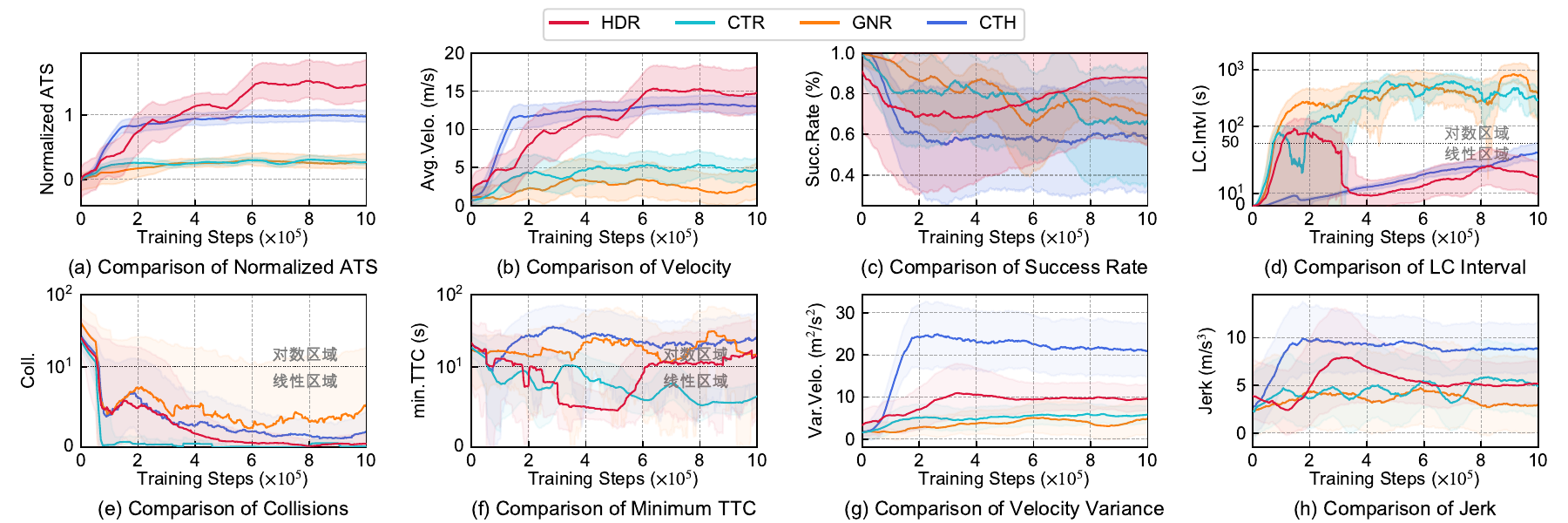}
	\caption{Performance of the MADDPG algorithm guided by different reward functions.}
	\label{fig_marl_rew_comp_MADDPG}
\end{figure*}

Fig. \ref{fig_marl_rew_comp_QMIX} shows the training return curves and key performance metrics for the QMIX algorithm under the four reward functions. Regarding convergence speed and stability, the normalized ATS curve for HDR exhibits the fastest initial climb and converges to the highest performance level after approximately $6 \times 10^5$ training steps. In contrast, GNR and CTR show significantly lower learning efficiency, with slow convergence speeds and poor final performance.

In terms of policy quality and balance, HDR not only achieves the best overall score but also demonstrates the best balance across various metrics in the learned policy. It achieves the highest average speed while maintaining the highest task success rate. Furthermore, it is the only method where the number of collisions converges rapidly to near zero, demonstrating superior safety.

Consistent with findings in the previous section, the performance of CTH is poor, with collision counts even rebounding in the later stages of training. This further confirms the conflict between the centering reward and the HDR mechanism; the baseline subtraction operation in CTR likely disrupts the key gradient signals provided by HDR.

Fig. \ref{fig_marl_rew_comp_MAPPO} presents the comparative results for the MAPPO algorithm. As a policy gradient method, MAPPO's performance is more sensitive to the quality of reward signals and the variance of gradient estimation. Regarding convergence speed and stability, the advantage of HDR is even more pronounced in MAPPO. The ATS curve for HDR is the only one capable of stably converging to a high performance level. The GNR, CTR, and CTH reward functions all failed to learn effective cooperative policies under the MAPPO framework, with their ATS values and average speeds remaining at extremely low levels and task success rates being highly unstable.

Regarding policy quality, the policy guided by HDR far surpasses other methods in safety and driving smoothness. This result indicates that for policy gradient algorithms, the sparse or high-variance signals provided by GNR or CTR are insufficient to support effective exploration in complex multi-agent game spaces. In contrast, the dense and physically meaningful differential signals provided by HDR are key to successful learning.

Fig. \ref{fig_marl_rew_comp_MADDPG} shows the results for the MADDPG algorithm. In terms of convergence speed and stability, compared to QMIX and MAPPO, MADDPG exhibits greater instability during training under all reward functions, particularly with severe fluctuations in task success rates. Nevertheless, HDR still demonstrates the fastest initial learning speed and the highest final ATS value.

Regarding policy quality, under the MADDPG framework, HDR is the only method capable of learning a policy that balances efficiency and safety. Policies guided by other reward functions are not only inefficient but also fail to effectively resolve collision issues. The collision frequency for CTH reaches as high as 0.145 collisions/hour, far exceeding other methods, indicating the worst safety performance.

In summary, whether under value decomposition-based, policy gradient-based, or Actor-Critic MARL frameworks, the Hybrid Differential Reward demonstrates consistent and significant superiority. The dense, high signal-to-noise ratio gradient signals it provides greatly enhance algorithm convergence speed and stability, enabling agents to learn high-quality cooperative policies that achieve an optimal balance among conflicting objectives such as safety, efficiency, and task completion rates.

\subsubsection{Validation of Effectiveness on Online Planning (MCTS)}

This subsection selects the Monte Carlo Tree Search algorithm as a representative of online planning methods to verify whether the HDR signal can provide more effective decision guidance. MCTS evaluates the long-term value of actions through forward simulation, and its search efficiency heavily relies on the quality of reward signals during the simulation process. Therefore, this experiment compares the impact of different reward functions on the quality of cooperative policies derived by the MCTS algorithm under varying computational budgets.

Specifically, this section compares the performance of the MCTS algorithm guided by four reward functions: MCTS+HDR, MCTS+CTR, MCTS+GNR, and MCTS+CTH. The experiment is conducted in a specific interaction scenario with a fixed number of agents, including several CAVs and HDVs. This setting is used to evaluate the algorithm's performance in solving a defined, non-continuous flow game problem.

\begin{figure*}[t]
	\centering
	\includegraphics[width=1.0\linewidth]{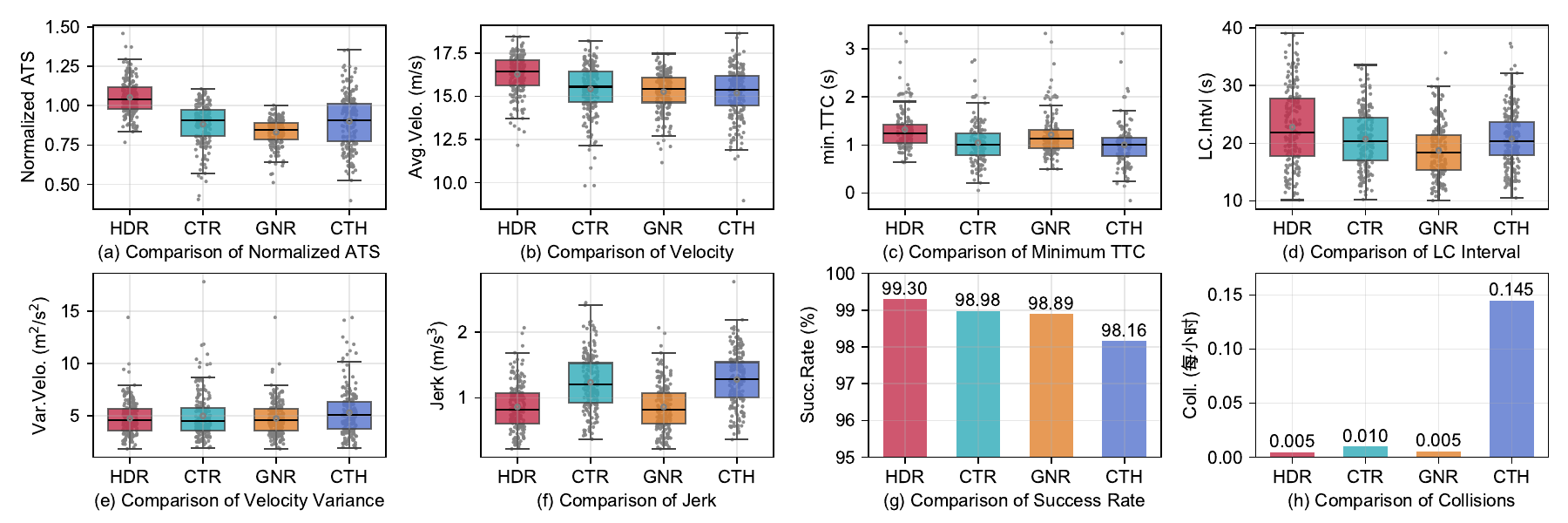}
	\caption{Comparison of objective metrics for the Monte Carlo Tree Search algorithm guided by different reward functions.}
	\label{fig:ch2_rew_mcts_metric}
\end{figure*}

Fig. \ref{fig:ch2_rew_mcts_metric} displays the performance comparison of MCTS policies guided by the four reward functions across various objective metrics under a fixed search budget. The results show that the MCTS+HDR method demonstrates significant advantages across almost all key performance dimensions. Specifically, HDR not only achieves the highest ATS value but also outperforms MCTS+GNR and MCTS+CTR in traffic efficiency (average speed), safety (average minimum TTC), task completion rate, and driving smoothness (lane change interval, speed variance, jerk). This result strongly proves that the HDR reward method can provide more precise and efficient heuristic information for online search algorithms, enabling them to converge to high-quality cooperative policies more quickly within limited simulations.

It is worth noting that the MCTS+CTH method performs poorly on several metrics, particularly being inferior to other methods in task completion rate and collision frequency. This is likely due to a principled conflict between the two mechanisms. The core idea of centering reward is to reduce value estimation variance by subtracting a dynamic reward baseline. However, the design goal of the Hybrid Differential Reward is precisely to provide an information-dense, non-zero-mean signal with clear guidance direction through potential function differences and action gradients. Applying CTR to HDR, the centering operation may disrupt the carefully designed gradient signals of HDR, interfering with its effective heuristic guidance for the search process, thereby leading to negative effects.

\begin{figure}[h]
	\centering
	\includegraphics[width=1.0\columnwidth]{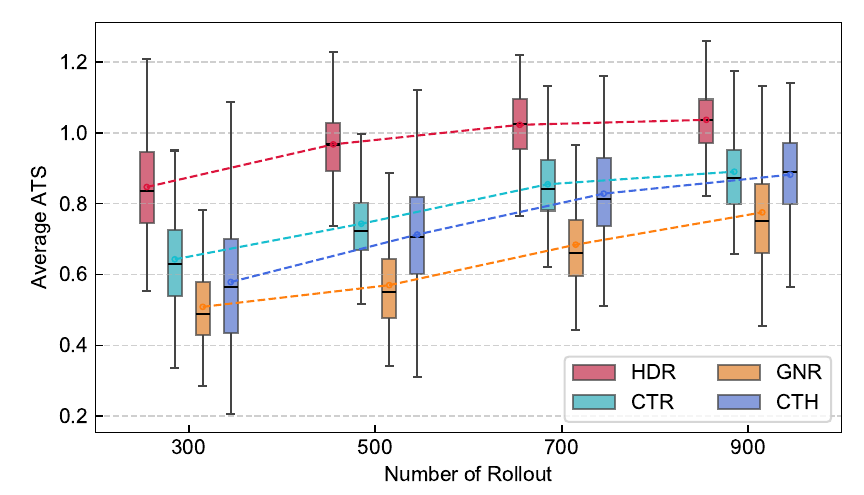}
	\caption{Requirements of different reward functions for the number of searches.}
	\label{fig:ch2_rew_mcts_asc_roll}
\end{figure}

Fig. \ref{fig:ch2_rew_mcts_asc_roll} further analyzes the dependency of the algorithm's comprehensive performance on the number of searches. The experimental results indicate that the MCTS+HDR method consistently maintains optimal performance across all tested search budget levels. More importantly, the marginal growth of the HDR curve has tended to flatten, indicating that it approaches convergence at a lower number of searches. In contrast, the performance curves of other methods still show significant upward trends. This suggests that to achieve the same performance level as HDR, other reward functions would require a computational budget far exceeding that of HDR. In conclusion, HDR not only improves the final quality of the policy but also significantly enhances the search efficiency of the MCTS algorithm.

\section{Conclusion}
\label{sec_conclusion}

This paper addresses a critical challenge in multi-vehicle cooperative driving characterized by high-frequency continuous decision-making: the problem of vanishing reward differences. To resolve this, we propose a novel and efficient Hybrid Differential Reward (HDR) mechanism.

First, through theoretical analysis and visualization, we formally elucidate the issue of low signal-to-noise ratio (SNR) in policy gradients associated with traditional state-based reward functions under high-frequency decision-making. We attribute this issue to the quasi-steady nature of traffic states and the physical proximity of actions. Second, we propose the HDR framework, which innovatively integrates two complementary differential signals:
\begin{enumerate}
	\item \textbf{Temporal Difference Reward (TRD):} Defined based on a potential function, it ensures optimal policy invariance and provides the correct direction for long-term optimization.
	\item \textbf{Action Gradient Reward (ARG):} By directly measuring the marginal utility of actions, it significantly improves the SNR of local policy gradients, resolving the issue of low action differentiation.
\end{enumerate}
Subsequently, we formulate the multi-vehicle cooperative driving task as a Multi-Agent Partially Observable Markov Game (POMDPG) with a time-varying agent set. We also provide a complete instantiation and a computable derivation scheme for the HDR mechanism within this complex scenario.

Finally, through extensive validation in a unified simulation environment using both online planning (MCTS) and offline learning (QMIX, MAPPO, MADDPG) algorithms, we demonstrate that the HDR mechanism achieves the fastest convergence speed and the highest final performance across all tested frameworks. The results indicate that HDR effectively guides agents to learn cooperative driving policies that strike an optimal balance among safety, efficiency, and driving smoothness.

The HDR mechanism proposed in this paper provides a general and effective paradigm for addressing reward design challenges in high-frequency continuous control. Building on this, future work can be explored in the following directions:

1) \textbf{Adaptive Hyperparameter Adjustment:} The mixing weight $\alpha$ in this paper is currently a manually set hyperparameter. Future research could explore dynamically adjusting $\alpha$ based on environmental complexity or learning stages to further enhance model adaptability and learning efficiency.

2) \textbf{Generalization and Cross-Domain Application:} Future work may extend the HDR framework to other complex multi-agent systems requiring high-frequency, continuous decision-making, such as multi-robot collaboration and UAV swarms, to further validate its theoretical and practical value.

3) \textbf{Integration with Causal Inference:} It is worthwhile to investigate the integration of the Action Gradient Reward (ARG) in HDR with causal inference methods in reinforcement learning (e.g., Counterfactual Reasoning) to provide more interpretable and robust assessments of action utility.


\bibliographystyle{IEEEtran.bst}
\bibliography{hdr_rew}

\end{document}